\documentclass[letterpaper, 10 pt, conference]{ieeeconf}  

\IEEEoverridecommandlockouts                              

\usepackage{graphics} 
\usepackage{epsfig} 
\usepackage{mathptmx} 
\usepackage{amsmath} 
\usepackage{amssymb,amsfonts}  
\usepackage{newtxmath}
\usepackage{multirow}
\usepackage{subcaption}
\usepackage{array,colortbl,multirow,multicol,booktabs,ctable}

\title{\LARGE \bf
SUMO-MCP: Leveraging the Model Context Protocol for \\ Autonomous Traffic Simulation and Optimization
}

\author{Chenglong Ye$^{\dag}$, Gang Xiong$^{\dag}$, Junyou Shang, \\ Xingyuan Dai, Xiaoyan Gong, Yisheng Lv$^{\ddag}$
\thanks{This paper was supported in part by the National Natural Science Foundation of China under Grants 62303462 and 62271485. Beijing Natural Science Foundation under grant L241016. Chongqing Transportation Technology Project (CQJT-CZKJ2024-04)}
\thanks{Chenglong Ye, Gang Xiong, Junyou Shang, Xingyuan Dai, Xiaoyan Gong, and Yisheng Lv are with the State Key Laboratory of Multimodal Artificial Intelligence Systems, Institute of Automation, Chinese Academy of Sciences, Beijing, 100190, China. Chenglong Ye, Junyou Shang, Xingyuan Dai, and Yisheng Lv are also with the School of Artificial Intelligence, University of Chinese Academy of Sciences, Beijing 100049, China.}
\thanks{$^{\dag}$Chenglong Ye and Gang Xiong are co-first authors.}
\thanks{$^{\ddag}$Yisheng Lv is corresponding author.}%
}

\begin{document}

\maketitle
\thispagestyle{empty}
\pagestyle{empty}


\begin{abstract}

Traffic simulation tools, such as SUMO, are essential for urban mobility research. However, such tools remain challenging for users due to complex manual workflows involving network download, demand generation, simulation setup, and result analysis. In this paper, we introduce SUMO-MCP, a novel platform that not only wraps SUMO’s core utilities into a unified tool suite but also provides additional auxiliary utilities for common preprocessing and postprocessing tasks. Using SUMO-MCP, users can issue simple natural-language prompts to generate traffic scenarios from OpenStreetMap data, create demand from origin–destination matrices or random patterns, run batch simulations with multiple signal-control strategies, perform comparative analyses with automated reporting, and detect congestion for signal-timing optimization. Furthermore, the platform allows flexible custom workflows by dynamically combining exposed SUMO tools without additional coding.  Experiments demonstrate that SUMO-MCP significantly makes traffic simulation more accessible and reliable for researchers. We will release code for SUMO-MCP at https://github.com/ycycycl/SUMO-MCP in the future.

\end{abstract}


\section{Introduction}

The emergence of large language models (LLMs) has had significant impacts on intelligent transportation systems. These advanced models can understand plain-language instructions, analyze data, and perform complex tasks. When equipped with the ability to call external tools, an LLM can act as an agent—an intelligent program that understands user requests and automatically carries out complex tasks. Such agents offer unprecedented opportunities for building smarter, more efficient urban mobility systems.

These agents rely on specialized transportation tools to perform their tasks effectively.  One key tool is microscopic traffic simulation software, such as SUMO (Simulation of Urban Mobility), widely used for evaluating traffic scenarios and optimizing signal timings. However, despite its popularity, SUMO remains difficult for non-experts to use, largely due to its complex interfaces, especially the Traffic Control Interface (TraCI), which requires significant expertise.

Recent studies have explored the integration of large language models (LLMs) with SUMO to simplify user interactions. However, these methods still require users to follow fixed and predefined steps. Users cannot easily change or adjust these steps when encountering new tasks or unexpected requirements. Moreover, these systems depend on custom-built interfaces, making it difficult to integrate additional tools or handle flexible simulation workflows.

To solve this problem, this paper presents SUMO-MCP, a prompt-assisted platform that makes SUMO “chat-ready.”  By wrapping SUMO’s core utilities into an MCP-compatible Tool Suite and adding key Auxiliary Utilities, SUMO-MCP lets users type a single natural-language request—e.g., “Simulate evening peak in Haidian, compare Webster vs.\ GreenWave”—and have an agent automatically select, load, and run the necessary tools in order.  Behind the scenes, Model Context Protocol (MCP)’s client–server protocol provides dynamic discovery, on-demand loading, and structured error handling, turning hours of scripting into seconds of prompting. This approach makes SUMO significantly easier and more intuitive to use, especially for users who lack extensive programming or traffic simulation expertise.

\begin{figure*}[t]
    \centering
    \begin{subfigure}[b]{0.46\textwidth}
        \centering
        \includegraphics[width=\linewidth]{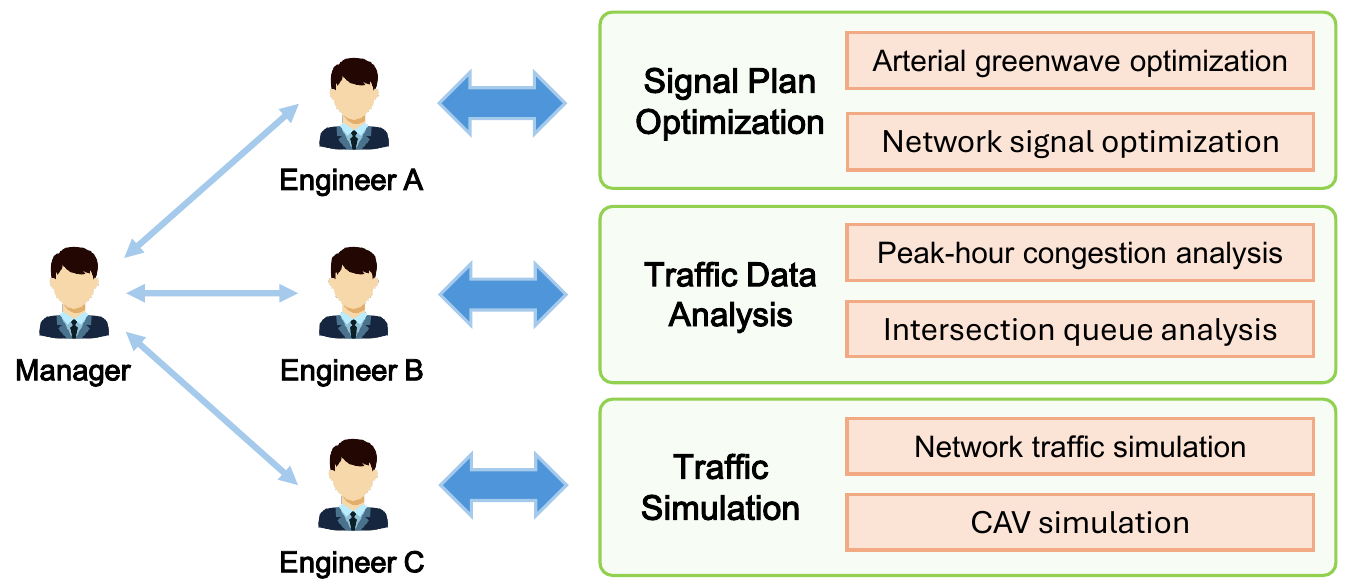}
        \caption{Traditional workflow}
    \end{subfigure}
    \hfill
    \begin{subfigure}[b]{0.46\textwidth}
        \centering
        \includegraphics[width=\linewidth]{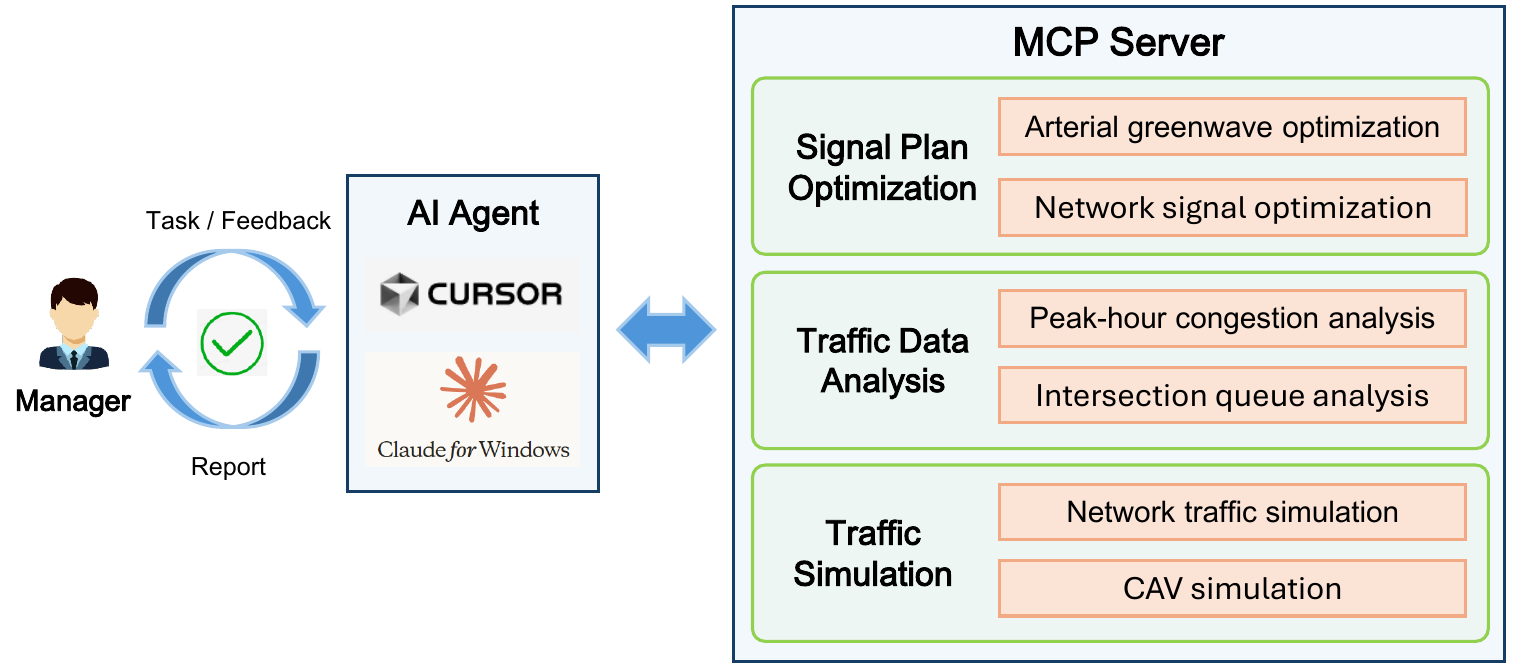}
        \caption{Agent-driven workflow}
    \end{subfigure}
    \caption{Comparison of traffic-management workflows. The proposed agent–MCP approach (b) replaces manual multi-engineer coordination (a) with a single agent that dynamically discovers and orchestrates the required tools.}
    \label{fig:workflow_comparison}
\end{figure*}

Fig. \ref{fig:workflow_comparison} compares the traditional manual traffic management workflow with our agent-driven framework. In the traditional workflow, engineers must manually configure simulations and repeatedly analyze results. In contrast, our MCP-enabled agent automatically selects appropriate simulation tools, organizes the workflow dynamically, and handles traffic optimization tasks with minimal human intervention. 

The main contributions of this paper are as follows:
\begin{enumerate}
    \item We integrate MCP with SUMO for the first time and create an MCP-compatible SUMO Tool Suite, allowing agents to dynamically find and call SUMO tools;
    \item We develop a prompt-assisted Simulation Generation and Evaluation Tool, enabling users to easily set up, run, and compare multiple traffic-signal strategies;
    \item We implement a prompt-assisted Signal Control Optimization Tool, automatically detects traffic congestion and optimizes signal plans based on simulation results.
\end{enumerate}

SUMO-MCP demonstrates that prompt-assisted traffic simulation is not only feasible but also fast, flexible, and accessible to a broad range of users.  

\section{Related Work}
\subsection{SUMO and Traffic Simulation Automation}
SUMO is an open-source microscopic traffic simulator that supports multi-modal entities, handles large road networks, and ships with tools for network import, demand generation, signal control, and output analysis. Owing to its advantages, many traffic signal control relies on SUMO to implement and test new algorithms. Early TSC algorithms, such as IntelliLight \cite{wei2018intellilight}, CoLight \cite{wei2019colight}, SPTO \cite{yang2023semi}, TrafficWise \cite{hu2025trafficwise}, PLANT \cite{zhao2023parallel}, and DQNOP \cite{cheng2024deep}, apply deep reinforcement learning to traffic signal control tasks and verify the proposed RL algorithms via SUMO. The recent rise of LLMs (large language models) has opened new directions for TSC:  works such as LLMLight \cite{lai2023llmlight}, CoLLMLight \cite{yuan2025collmlight}, LLM-assisted light \cite{wang2024llm}, CityLight \cite{zeng2024citylight}, LLMAATSC \cite{tang2024large}, and LLMDUTSC \cite{tang2024large1} also propose LLM-assisted frameworks for traffic-signal management and integrate SUMO for traffic simulation and algorithm validation.

In addition to TSC, many autonomous driving-related studies have also used SUMO for simulation verification, such as SMDT \cite{wang2024smart}, BlaFT \cite{park2023blame}, HPTSim \cite{shi2024hptsim}, DLIO \cite{wang2024iterative}, ASC-HBMP \cite{zong2023human}, and RoboCar \cite{testouri2025robocar}.

These studies confirm SUMO’s versatility, yet they also reveal a key drawback: every SUMO simulation step must still be performed manually, resulting in a fragmented workflow that is both error-prone and time-consuming.

Recent work has begun to link large-language models (LLMs) with SUMO so that users can run simulations by talking to the system. For example, ChatSUMO  \cite{li2024chatsumo} lets a user type a request, then the LLM launches Python scripts that build a network and run a basic SUMO simulation. Open-TI  \cite{da2024open} offers a similar chat interface: a user asks a question, and the LLM starts the required SUMO runs in the background.

While ChatSUMO and Open-TI prove that SUMO can be controlled with plain-language requests, they still follow a rigid recipe. Each step is pre-programmed, so the LLM only plugs in parameters and cannot change the sequence when the task changes. In addition, these systems expose only a small set of hand-wrapped commands. Because there is no shared tool catalogue, the LLM cannot discover new SUMO utilities or combine them in new ways during run-time.

\subsection{MCP}
The Model Context Protocol (MCP) \cite{AnthropicMCP2024}, released by Anthropic in 2024, is designed specifically for agents. It gives agents a simple, standard way to call external tools through lightweight JSON-RPC messages. Each tool publishes a short, machine-readable description of its inputs and outputs, so the agent can discover what is available, choose the tools it needs, and invoke them in the proper order, without any custom APIs or hard-coded scripts. In this way, MCP lets an agent build and modify entire workflows on its own, greatly reducing integration effort and increasing flexibility.

Recent developments in agent-based AI frameworks such as AutoGen \cite{wu2024autogen}, ChatDev \cite{qian2024chatdev}, MetaGPT \cite{hong2024metagpt}, and CrewAI \cite{duan2024exploration} underscore the rising importance of flexible, schema-driven interfaces. These frameworks leverage agents' ability to dynamically orchestrate multi-step tasks, adaptively select appropriate tools based on runtime context, and autonomously handle errors through structured feedback.

Previous attempts to link LLMs with SUMO, such as ChatSUMO and Open-TI, depend on fixed, custom APIs, leaving little room for dynamic adaptation. To our knowledge, ours is the first work to pair SUMO with the MCP. By exposing the full SUMO toolchain as MCP services, an agent can discover, invoke, and recombine simulation tasks on demand, eliminating manual scripts and substantially improving flexibility and accessibility in traffic-simulation workflows.

\section{Methodology}
\subsection{System Overview}
\begin{figure*}[htbp] 
    \centering 
    \includegraphics[width=0.98\textwidth]{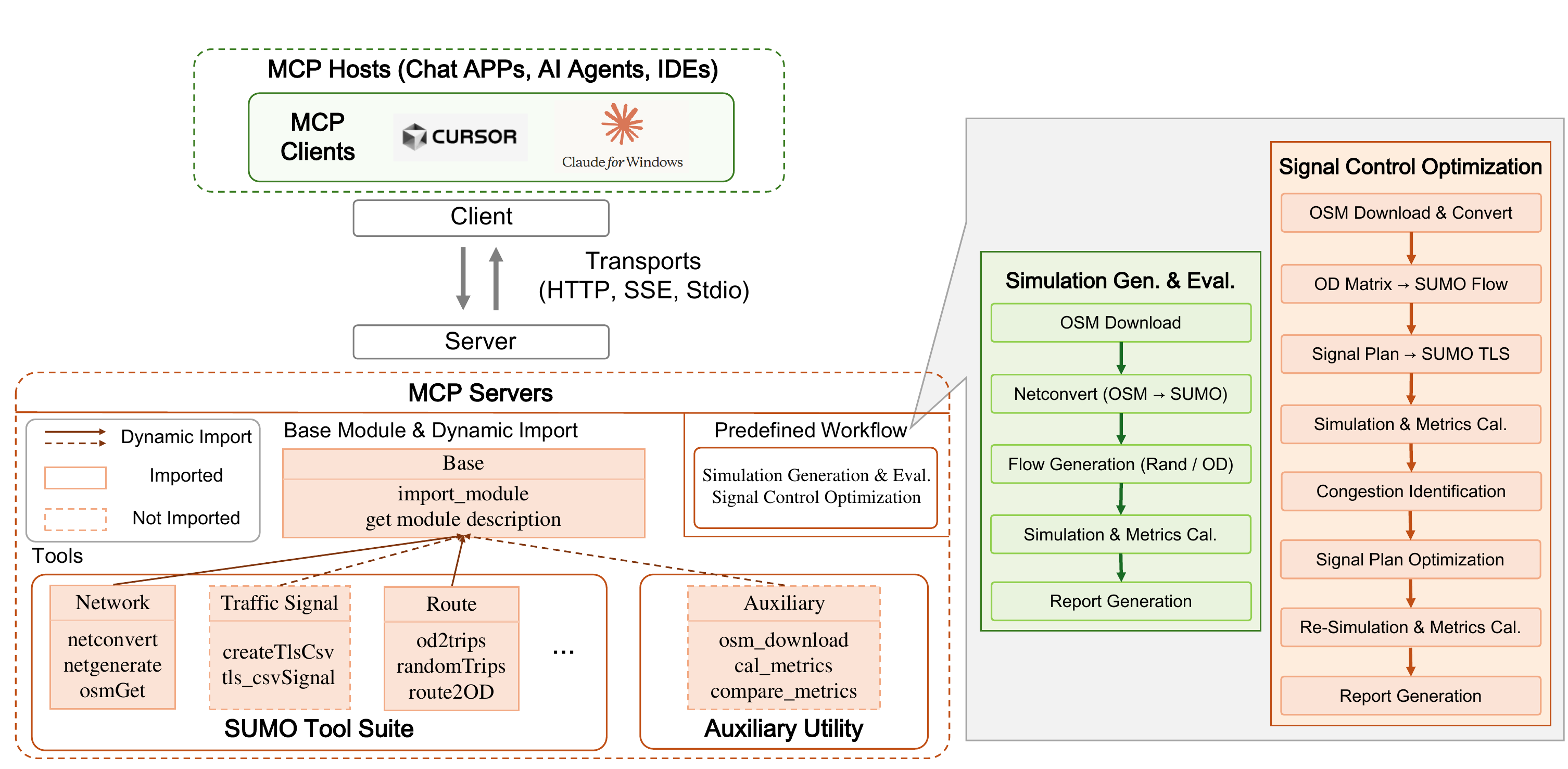} 
    \caption{Overall framework of SUMO-MCP} 
    \label{fig: Framework} 
\end{figure*}

SUMO-MCP turns plain-language prompts into complete traffic-simulation workflows with minimal effort.  As shown in Fig.~\ref{fig: Framework}, a conversational agent on the client side sends user requests to the MCP server.  The server, built on FastMCP, dynamically registers and runs only the needed tools—SUMO utilities and other auxiliary utilities—via a lightweight JSON-RPC interface.  We provide two prompt-assisted predefined workflows that guide users through common tasks without any manual coding: Simulation Generation \& Evaluation, and Signal Control Optimization (will be discussed below).

Specifically, SUMO-MCP supports the following tasks:
\begin{itemize}
    \item Automatic network downloading and conversion from OpenStreetMap;
    \item Traffic-demand generation from origin–destination (OD) matrices or random patterns;
    \item Batch simulation with multiple signal-control strategies (Fixed-Time, Actuated, Webster, GreenWave);
    \item Comparative analysis and reporting of key performance metrics;
    \item Detection of congested intersections and automatic optimisation of signal timings;
    \item \textbf{Custom workflows:} the LLM agent can mix and match any exposed SUMO tool to meet new user requests without extra coding.
\end{itemize}

\subsection{SUMO-MCP Client}
The SUMO-MCP Client is implemented as an LLM-based agent, which serves as an intelligent interface to interpret user requests and translates a user’s natural-language prompt into MCP calls. Upon receiving a natural-language prompt (e.g., ``Compare Webster and GreenWave control for Shanghai, China''), the agent first calls the server's $get\_module\_description$ method to get a list of available tool modules and their functionalities. Then, by calling $import\_module$, the agent dynamically imports only the specific tools it requires (such as $netconvert$, $randomTrips$, or $compare\_metrics$) to construct the simulation workflow. This selective import mechanism enables efficient, streamlined interactions without manual scripting.

\subsection{MCP Server: Base Module \& Dynamic Import}
The Base Module implements Dynamic Import, a lightweight mechanism that loads entire sub-modules only when the client explicitly requests them.
A \textbf{sub-module} is a small collection of related SUMO tools that serve a common purpose; for example, the \textit{Network} sub-module bundles \textit{netconvert}, \textit{netgenerate}, etc, whereas the \textit{Route} sub-module groups \textit{randomTrips}, \textit{od2trips}, etc.
The interaction proceeds in three steps:

When a prompt arrives, the client inspects available sub-modules, imports the selected ones, and the server executes each tool in turn.  This prompt-assisted workflow—from “download map” to “optimize signals”—runs end-to-end without manual intervention.

\begin{enumerate}
\item \textbf{Intent Analysis.}
After parsing a user’s natural-language request, the SUMO-MCP Client determines which sub-modules are needed.
\item \textbf{Tool Discover.}
Client calls $get\_module\_description$, then the server returns a catalogue of available sub-modules (Network, Route, Auxiliary, etc).
\item \textbf{Dynamic Import.}
The client calls $import\_module$ to import the required sub-modules. The Base Module then imports the corresponding sub-module.
\end{enumerate}

Because sub-modules are loaded only when they are needed, Dynamic Import keeps the server lightweight, shortens start-up time, and avoids the “tool overload” problem that can confuse the agents when too many tools are pre-loaded. This design also lets developers extend the tool suite incrementally without restarting the server or modifying the client.

\subsection{MCP Server: Tools}
The tool suite is organized into thematic sub-modules. Each sub-module remains inactive (as a lightweight placeholder) until explicitly imported by the agent. Upon import, FastMCP activates each SUMO utility in the sub-module, exposing it as a stateless Python function using the $@mcp.tool()$. 

\subsubsection{SUMO Tool Suite}
SUMO provides a wide range of command-line utilities such as $netconvert$, $netgenerate$, $randomTrips$, $duaRouter$, and many others. However, packaging all these tools into a single MCP server would cause unnecessary memory usage, slow startup times, and overwhelm the agent with many irrelevant tools. To solve this problem, we group related tools into functional sub-modules and load them only when required. The SUMO Tool Suite currently exposes nine functional sub-modules—network, route, traffic\_signal, detector, district, turn\_defs, visualization, xml, and import\_tool—each loaded on demand.

\subsubsection{Auxiliary Utility}
In addition, we provide auxiliary tools such as $osm\_download$, $run\_simulation$, $cal\_metrics$, $compare\_metrics$, and $tls\_to\_csv$ that cover routine steps in network acquisition, configuration, execution, and post processing. Each auxiliary tool is exposed in the same way as the SUMO tools above.

When the agent receives a user request, it first calls $get\_module\_description$ to obtain a machine-readable catalogue of available modules. It then selects the relevant subset according to user request and calls $import\_modules$ to register only the chosen tools with the FastMCP server. 

\subsection{MCP Server: Predefined Workflow}
To guide non-expert users, we provide two predefined workflows—Simulation Generation \& Evaluation and Signal Control Optimization—illustrated in Fig.~\ref{fig: Framework}.

\subsubsection{Simulation Generation \& Evaluation}
Given a natural-language prompt (e.g., “Compare different signal plans for Haidian district, Beijing during the evening peak”), the agent automatically orchestrates a complete SUMO study without requiring specialized knowledge. The agent sequentially acquires OpenStreetMap data, converts it into a SUMO-compatible network, generates vehicle demand (either randomly or from an OD matrix), executes batch simulations for four signal-control methods (Fixed-Time, Actuated, Webster, GreenWave), and generates a comparative performance report including average delay, travel time, and queue statistics.

\subsubsection{Signal Control Optimization}
Users provide only an OD matrix and initial signal timings in CSV format, without any SUMO-specific configuration. The agent first generates vehicle routes, then simulates baseline signal performance. After identifying congestion hotspots, it autonomously optimizes cycle length, green splits, and offsets for selected intersections, re-simulates the optimized scenario, and produces a final report summarizing signal adjustments and performance improvements.

Because every SUMO detail is hidden behind this single natural-language template, even non-experts can complete a SUMO simulation study in minutes. The workflow itself is only a prompt, not a rigid script, so advanced users can override any step, allowing the agent to reorder or streamline the workflow for custom analyses.

\subsection{Overall Workflow}
\begin{figure} 
    \centering 
    \includegraphics[width=0.98\linewidth]{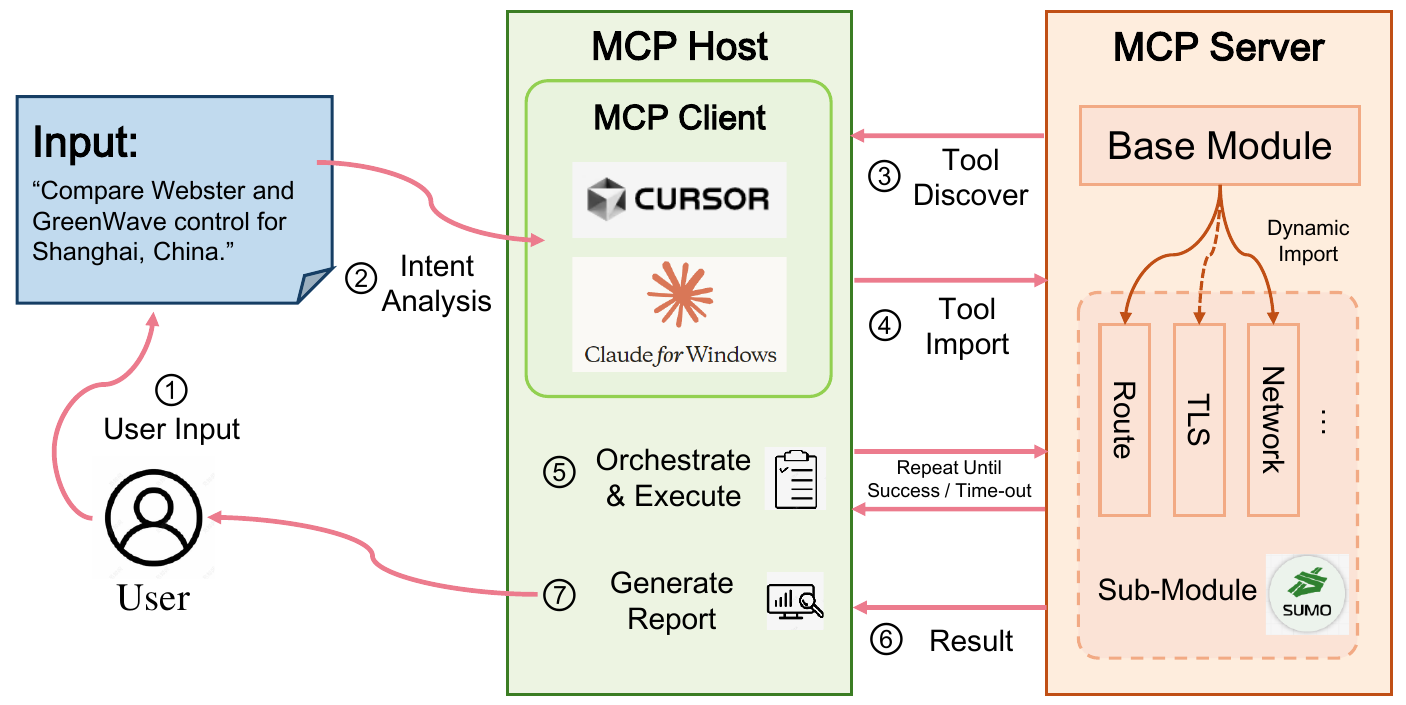} 
    \caption{The workflow of SUMO-MCP} 
    \label{fig:Workflow} 
\end{figure}
Fig. \ref{fig:Workflow} illustrates the run‑time sequence from user request to final report.  The workflow is agent‑centred and relies on MCP’s discovery–import mechanism to keep the server lightweight:

\begin{enumerate}
    \item \textbf{User Input}. The user issues a natural‑language request such as “Compare Webster and GreenWave control for Shanghai, China.”
    \item \textbf{Intent Analysis}. The agent parses the request and derives a high‑level plan (network acquisition, demand generation, dual‑strategy simulation, metric comparison).
    \item \textbf{Tool Discover ($get\_module\_description$)}. The agent queries the MCP server to obtain a machine‑readable catalogue of sub‑modules (e.g., Network, Route, Traffic Signal).
    \item \textbf{Tool Import ($import\_module$)}. Only the sub‑modules required for the plan are dynamically loaded; others remain placeholders. The import call is handled by the Base Module on the server.
    \item \textbf{Orchestrate \& Execute (loop)}. The agent invokes the imported tools in sequence and executes them; if a structured error is returned, it adjusts parameters and retries until success or time‑out.
    \item \textbf{Result Obtaining}. The agent obtains execution results from the MCP Server.
    \item \textbf{Report Generation}. After all steps succeed, the agent aggregates the results and generates a concise report for the user.
\end{enumerate}

By loading tools on demand and exploiting MCP’s structured errors, the agent builds an end‑to‑end SUMO workflow without hard‑coded scripts and can extend the pipeline at run‑time.

\begin{figure}
    \centering
    \includegraphics[width=1.0\linewidth]{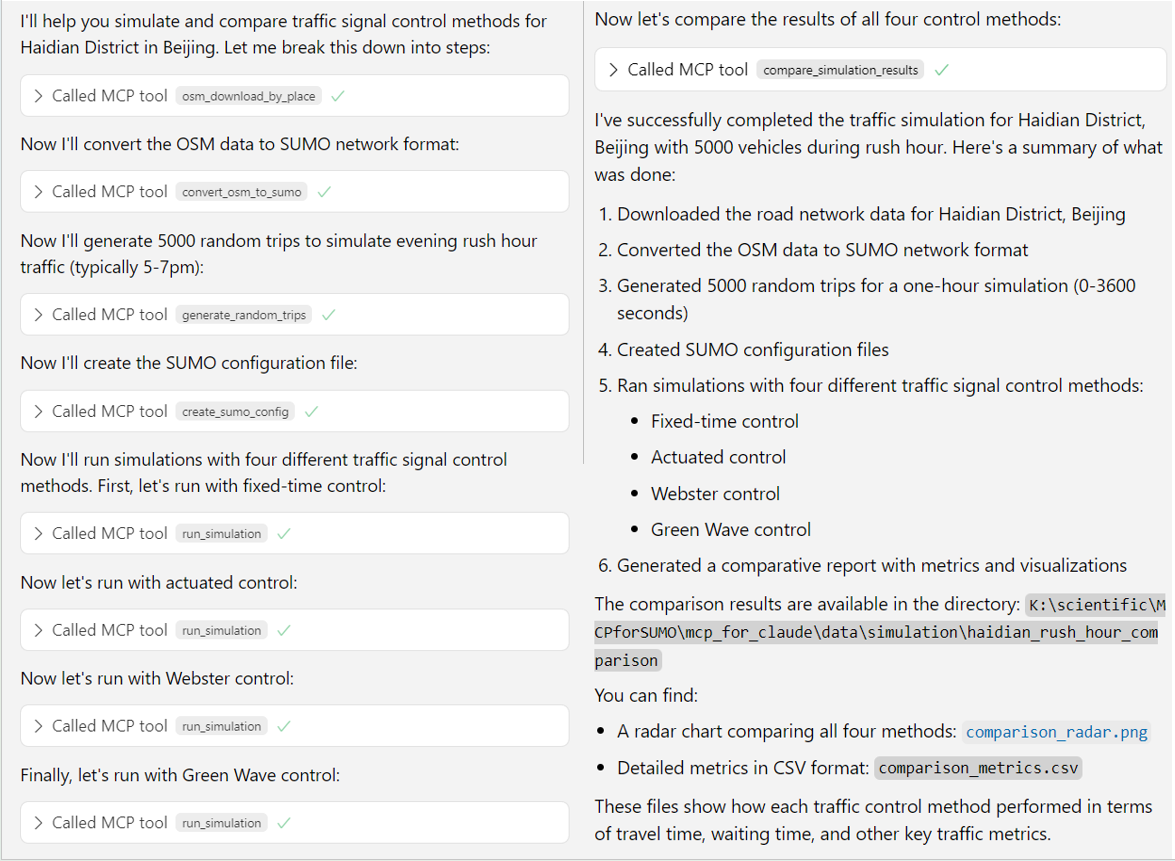}
    \caption{Process of Simulation Generation \& Evaluation}
    \label{fig:E1}
\end{figure}

\section{Experiment Result}
Our server runs on Python 3.10 with FastMCP 2.3.1, and SUMO v1.21.0. Each SUMO or auxiliary utility is wrapped as a small @mcp.tool() function that launches the relevant executable. Editors such as Cursor and Claude Desktop ship a native MCP‑client panel; users paste the server URL and can immediately call tools, so the SUMO‑MCP stack feels like a plug‑in installation.

The remainder of this section presents four experiments that evaluate functionality, Optimization benefit, dynamic‑import overhead, and MCP usability.

\subsection{Simulation Generation \& Evaluation}
We evaluate the Simulation Generation \& Evaluation tool using a natural-language prompt: \textbf{Download the road network in Chaoyang District, Beijing, simulate the evening rush hour traffic flow (5000), generate a one-hour simulation, evaluate the control effect of four signal control methods, and generate a comparative report.}

As shown in Fig. \ref{fig:E1}, the agent automatically decomposes this request into multiple steps: downloading and converting the road network, generating trips, running SUMO simulations for four signal control strategies, and generating a comparative report. Table \ref{tab:E1_result} shows the comparison result.

All steps are performed via MCP tool calls, with no manual configuration or scripting required.
This experiment demonstrates that complex multi-stage traffic simulation workflows can be fully automated through a single user instruction.
\begin{table}[htbp]
\caption{Performance Comparison of Four Signal-Control Methods}
\begin{center}
\scriptsize
\begin{tabular}{|c|c|c|c|}
\hline
\textbf{Method} & \textbf{Avg Travel Time} & \textbf{Avg Waiting Time} & \textbf{Avg Delay} \\
\cline{2-4}
 & (s) & (s/veh) & (s/veh) \\
\hline
Fixed     & 845.34 & 256.06 & 370.36 \\
Actuated  & 773.97 & 137.77 & 267.96 \\
Webster   & \textbf{703.13} & \textbf{77.76} & \textbf{179.38} \\
Green Wave& 838.35 & 246.62 & 360.05 \\
\hline
\end{tabular}
\label{tab:E1_result}
\end{center}
\end{table}

\subsection{Signal Control Optimization}
\begin{figure}
    \centering
    \includegraphics[width=1\linewidth]{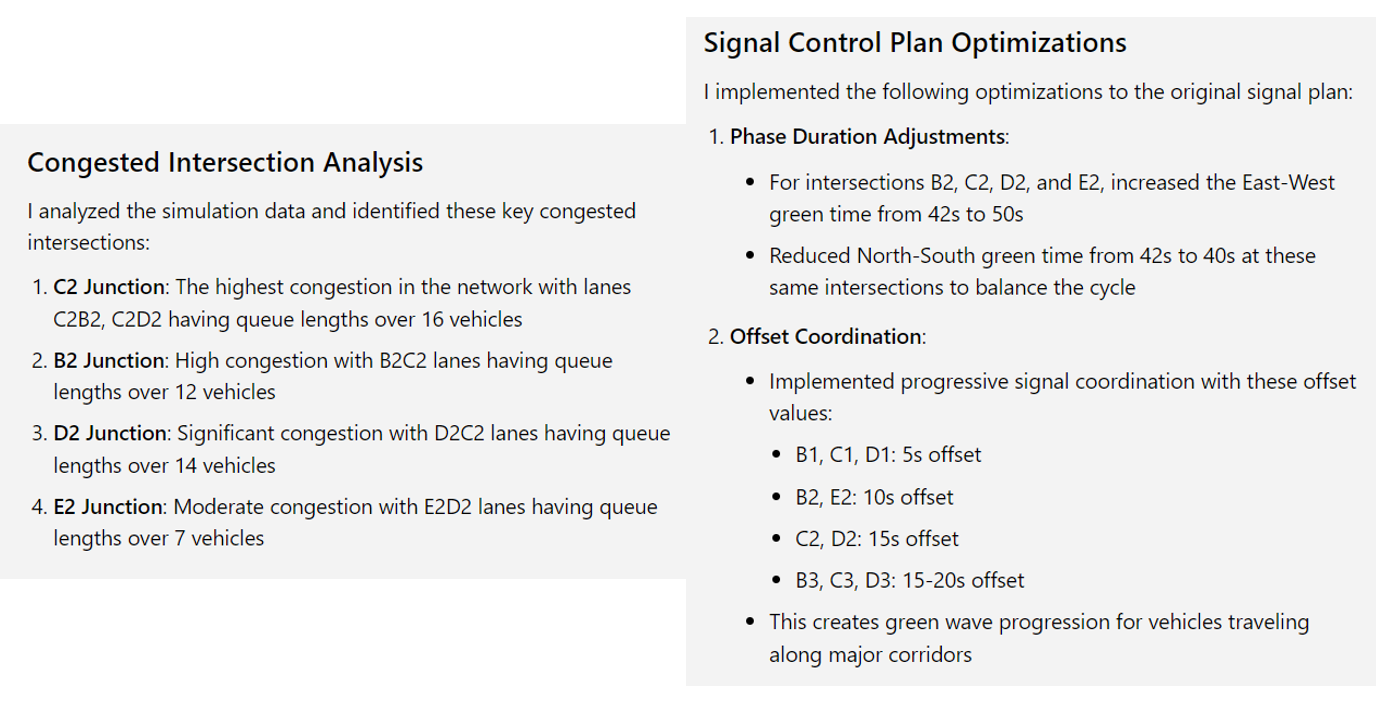}
    \caption{Agent-Driven Analysis and Optimization of Congested Intersections}
    \label{fig:E2_opt}
\end{figure}

We conduct the signal optimization experiment on a 5×5 intersection network from Xiong’an, China. The agent first analyzes the baseline simulation results, automatically identifying the most congested intersections (shown in Fig. \ref{fig:E2_opt}). It then optimizes durations and offset coordination for these key locations, producing an optimized signal plan.

Re-simulation with the optimized plan lead to significant reductions in vehicle delay and queue length, particularly at critical intersections. Table \ref{tab:key-metrics} reports the percentage improvement in key performance metrics, both network-wide and for the most congested junctions. These results demonstrate that our SUMO-MCP not only locates bottlenecks but also refines signal plans automatically, yielding tangible benefits without manual intervention or heuristic tuning.

\begin{table}[htbp]
\caption{Performance Improvement at Key Intersections Before and After Optimization}
\begin{center}
\begin{tabular}{|c|c|c|}
\hline
\textbf{Category} & \textbf{Metric} & \textbf{Improvement (\%)} \\
\hline
\multicolumn{3}{|l|}{\textbf{Overall Metrics}} \\
\hline
 & Average Travel Time & 2.30 \\
 & Average Waiting Time & 3.96 \\
 & Average Delay & 6.86 \\
\hline
\multicolumn{3}{|l|}{\textbf{Key Intersection Metrics}} \\
\hline
C2 & Queue Time \& Length & 21.45 \& 16.85 \\
E2 & Queue Time \& Length & 11.34 \& 7.59 \\
D2 & Queue Time \& Length & 6.55 \& 7.05 \\
B2 & Queue Time \& Length & 8.24 \& 7.72 \\
\hline
\end{tabular}
\label{tab:key-metrics}
\end{center}
\end{table}

\subsection{Dynamic-Import Ablation}

\begin{table}[htbp]
\caption{Task Completion Time and Memory Usage: Dynamic Import vs. Pre-load}
\begin{center}
\begin{tabular}{|c|c|c|}
\hline
\textbf{Region} & \textbf{Dynamic Import} & \textbf{Pre-load All} \\
\hline
Fengtai, Beijing & \textbf{81\,s} / 268\,MB   & 130\,s / 267\,MB \\
Pudong, Shanghai  & \textbf{93\,s} / \textbf{500\,MB}   & 151\,s / 525\,MB \\
Tianhe, Guangzhou  & \textbf{70\,s} / 272\,MB  & 86\,s / 272\,MB \\
Guangming, Shenzhen  & \textbf{55\,s} / 220\,MB & 160\,s / 221\,MB \\
Jiaxing & \textbf{183\,s} / 674\,MB & 313\,s / 665\,MB \\
\hline
\end{tabular}
\label{tab:dynamic-vs-preload}
\end{center}
\end{table}

We evaluate the impact of dynamic module import(see Methodology §III.C for details) versus pre-loading all tools by executing five representative network download and simulation tasks on both configurations. As shown in Table \ref{tab:dynamic-vs-preload}, dynamic import reduces total completion time for every scenario, while peak memory usage remains similar.

Beyond raw efficiency, we observe a notable qualitative difference: when all modules are pre-loaded, the agent is more prone to “tool overload”—sometimes overlooking available tools and instead reverting to manual command-line invocation. This results in repeated execution errors and prolonged recovery times. Furthermore, as multiple errors accumulated and the conversational context length increased, the agent occasionally forgot its established workflow and began to take actions outside the intended plan.

These findings highlight that dynamic import not only improves efficiency but also helps maintain agent focus, reduces error cascades, and supports more robust autonomous task execution in complex multi-tool environments.

\subsection{MCP vs Direct CLI Ablation}

\begin{table}[htbp]
\caption{MCP vs Direct CLI: Task Time and Tool Invocation Count}
\begin{center}
\begin{tabular}{|c|c|c|c|c|}
\hline
\textbf{Region} & \multicolumn{2}{|c|}{\textbf{MCP}} & \multicolumn{2}{|c|}{\textbf{Direct CLI}} \\
\cline{2-5}
 & \textbf{Time (s)} & \textbf{Calls} & \textbf{Time (s)} & \textbf{Calls} \\
\hline
Beijing Fengtai    & \textbf{63} & \textbf{6}   & 100 & 13 \\
Shanghai Pudong    & \textbf{78} & \textbf{3}   & 111 & 10 \\
Guangzhou Tianhe   & \textbf{72} & \textbf{4}   & 115 & 16 \\
Shenzhen Guangming & \textbf{62} & \textbf{6}   & 100 & 14 \\
Jiaxing            & \textbf{95} & \textbf{6}   & 102 & 10 \\
\hline
\textbf{Average}   & \textbf{74} & \textbf{5.0} & 106 & 12.6 \\
\hline
\end{tabular}
\label{tab:mcp-vs-cli}
\end{center}
\end{table}

To assess the practical impact of the MCP interface, we compare agent performance on a typical scenario (“convert network and generate simulation”) using either the standardized MCP tool calls or direct command-line execution.

As shown in Table \ref{tab:mcp-vs-cli}, the MCP approach requires fewer tool calls and less execution time across all tested regions. In contrast, direct CLI scripting is significantly more error-prone, as parameter mismatches frequently lead to task retries. In several cases, the workflow stalls entirely, necessitating manual intervention to resume. These findings highlight that, beyond improving efficiency, MCP’s schema-driven interface substantially reduces the likelihood of parameter errors and workflow interruptions. The results demonstrate the robustness and usability advantage of MCP-enabled automation, especially for complex multi-step simulation tasks.

\section{Conclusion and Future Work}

This paper presents SUMO-MCP, a prompt-assisted platform that connects SUMO to LLM agents via the MCP, significantly reducing the technical barrier for orchestrating complex simulation workflows. Through a series of experiments, we demonstrate that the system enables automated scenario generation, signal optimization, and robust, efficient tool management with minimal manual intervention. The results show that this prompt-first design slashes setup time, reduces tool calls, and virtually eliminates common scripting errors, making complex traffic studies accessible to all users.

Looking ahead, we see several directions for further improvement. First, MCP could be extended to support dynamic registration and discovery of file-based resources generated during simulation, allowing the agent to track, reference, and manage intermediate artifacts as “Resources” in the MCP registry. This approach would fundamentally resolve path management and file-handling errors. Second, future work could explore more advanced agent planning strategies and the development of user-friendly frontends for interactive experiment design. 

\addtolength{\textheight}{-12cm}   









\bibliographystyle{IEEEtran}
\bibliography{main}

\end{document}